\documentclass{article}
\usepackage{arxiv}

\usepackage[utf8]{inputenc} % allow utf-8 input
\usepackage[T1]{fontenc}    % use 8-bit T1 fonts
\usepackage{url}            % simple URL typesetting
\usepackage{booktabs}       % professional-quality tables
\usepackage{amsfonts}       % blackboard math symbols
\usepackage{nicefrac}       % compact symbols for 1/2, etc.
\usepackage{microtype}      % microtypography
\usepackage{graphicx,multirow,subfig}
\usepackage{hyperref}       % hyperlinks

\newcommand{\color}[1]{}

\title{Detecting soccer balls with reduced neural networks: a comparison of multiple architectures under constrained hardware scenarios
}

\author{
  Douglas De Rizzo Meneghetti\\
  Department of Electrical Engineering\\
  FEI University Center\\
  S\~ao Bernardo do Campo, SP 09850-901\\
  \texttt{douglasrizzo@fei.edu.br}\\
   \And
  Thiago Pedro Donadon Homem\\
  Federal Institute of Education, Science and Technology of S\~ao Paulo\\
  S\~ao Paulo, SP 05110-000 \\
	\texttt{thiagohomem@ifsp.edu.br}\\
   \And
  Jonas Henrique Renolfi de Oliveira \\
  Department of Electrical Engineering\\
  FEI University Center\\
  S\~ao Bernardo do Campo, SP 09850-901\\
  \texttt{jonashro@gmail.com}\\
   \And
  Isaac Jesus da Silva \\
  Department of Electrical Engineering\\
  FEI University Center\\
  S\~ao Bernardo do Campo, SP 09850-901\\
  \texttt{isaacjesus@fei.edu.br}\\
   \And
  Danilo Hernani Perico\\
  Department of Electrical Engineering\\
  FEI University Center\\
  S\~ao Bernardo do Campo, SP 09850-901\\
  \texttt{dperico@fei.edu.br}\\
   \And
  Reinaldo Augusto da Costa Bianchi \\
  Department of Electrical Engineering\\
  FEI University Center\\
  S\~ao Bernardo do Campo, SP 09850-901\\
  \texttt{rbianchi@fei.edu.br}\\
}

\date{Sep. 27, 2020}

\begin{document}

\maketitle

\begin{abstract}
	Object detection techniques that achieve state-of-the-art detection accuracy employ convolutional neural networks, implemented to have optimal performance in graphics processing units. Some hardware systems, such as mobile robots, operate under constrained hardware situations, but still benefit from object detection capabilities. Multiple network models have been proposed, achieving comparable accuracy with reduced architectures and leaner operations. Motivated by the need to create an object detection system for a soccer team of mobile robots, this work provides a comparative study of recent proposals of neural networks targeted towards constrained hardware environments, in the specific task of soccer ball detection. We train multiple open implementations of MobileNetV2 and MobileNetV3 models with different underlying architectures, as well as YOLOv3, TinyYOLOv3, YOLOv4 and TinyYOLOv4 in an annotated image data set captured using a mobile robot. We then report their mean average precision on a test data set and their inference times in videos of different resolutions, under constrained and unconstrained hardware configurations. Results show that MobileNetV3 models have a good trade-off between mAP and inference time in constrained scenarios only, while MobileNetV2 with high width multipliers are appropriate for server-side inference. YOLO models in their official implementations are not suitable for inference in CPUs.
	% \PACS{PACS code1 \and PACS code2 \and more} não existe mais classificação PACS
	% \subclass{MSC 68T07 \and MSC 68T40 \and MSC 68T45} só existem no MSC 2020 e não no 2010, que é o que o artigo pede
\end{abstract}

\section{Introduction}
\label{sec:intro}

% COISAS PARA LEMBRAR NESTE ARTIGO
% entre 12 a 24 páginas (FEITO)
% o título tem que ser diferente do anterior (FEITO)
% o conteúdo tem que ser 70% diferente do anterior
% as seções têm que ser as mesmas do anterior (seções estão corretas, subseções divergem porque colocamos mais coisa)
% deve haver um parágrafo na introdução descrevendo como esse artigo expande o anterior (FEITO)

The recent successes in the field of object detection are mostly due to the use of deep neural networks, more specifically convolutional neural networks (CNN). The underlying operations that compose CNNs are highly optimized for execution in graphics processing units (GPU). However, in some domains, GPUs may be unavailable and these processes must be executed in CPUs. One such domain is mobile robotics, in which there may be limitations regarding space, weight and energy consumption, constraining the robot's hardware to contain only CPUs, thus hindering the performance of systems based on deep learning techniques.

With these limitations in mind, this work presents a study of the performance of recent CNN architectures applied to object detection tasks and proposed for constrained hardware settings. We train these models in the task of soccer ball detection and compare their mean average precision (mAP) in a test data set, as well as their inference time in both constrained and unconstrained hardware settings. With this study, we aim to guide readers in their choice of neural network when implementing an object detection system for mobile robots.

%
% \change{inserir algumas references}
%

We expand preceding work~\cite{deOliveira2019} by analyzing the newer MobileNetV2~\cite{Sandler2018} and MobileNetV3~\cite{Howard2019a} models, with more combinations of the width multiplier and input resolution parameters, as well as by including the YOLOv3~\cite{Redmon2018} and YOLOv4~\cite{Bochkovskiy2020} object detection algorithms and their "tiny" counterparts. This work also presents inference time results in both constrained and unconstrained environments and for multiple resolutions of input videos, providing readers with more relevant, up-to-date and comprehensive results to make decisions regarding the choice of neural network model to use in a soccer ball detection system (or any other comparable single-object detection system) under local, mobile, constrained hardware scenarios as well as when remote and unconstrained hardware configurations may be available.

This work is organized as follows: section~\ref{sec:background} lists the recent advances in the state-of-the-art in object detection, as well as techniques to create smaller network topologies while still maintaining high detection accuracy. We also describe the network architectures utilized in this work and their detection mechanisms. Section~\ref{sec:related} depicts related works. The methodology used throughout the experiments is presented in section~\ref{sec:methodology}. Section~\ref{sec:experiments} presents the experiments and the results are discussed in section~\ref{sec:results}. Lastly, section \ref{sec:conclusions} provides the conclusions and future work.

\section{Research Background}
\label{sec:background}

In recent years, object detection techniques have advanced at great pace due to the equally fast advances of deep learning and convolutional neural networks applied to computer vision~\cite{Zhao2019a}. Two-stage detectors such as Faster R-CNN~\cite{Ren2015} first generate region proposals and then detect objects only in the selected regions, while one-stage detectors such as YOLO~\cite{Redmon2015} and SSD~\cite{Howard2017} generate bounding box coordinates and class predictions at the same time, with YOLO using only convolutional layers for this task.

More recently, effort has centered around building strategies for efficiently scaling network models, reaching a trade-off between FLOPS, number of trainable parameters and accuracy. The MobileNetV3 architecture~\cite{Howard2019a} has been partially achieved via hardware-aware neural architecture search techniques~\cite{Yang2018c,Tan2019}, while AmoebaNet's architecture~\cite{Real2019}, which achieved state-of-the-art classification accuracy on ImageNet~\cite{Deng2009}, was evolved using evolutionary algorithms.

Other techniques attempt to shrink or expand the dimensions of convolutional layers using hyperparameters. MobileNetV1~\cite{Howard2017} introduced the width multiplier and input resolution parameters, discussed later in the text, while EfficientNet~\cite{Tan2020a} and EfficientDet~\cite{Tan2020} use a compound coefficient to scale all three dimensions of convolutional layers in order to maximize the network's accuracy. This, allied with neural architecture search, introduced the current state-of-the-art in image classification and object detection using CNNs.

\subsection{MobileNets}

Introduced in~\cite{Howard2017}, MobileNets are convolutional neural network architectures whose number of trainable parameters can be controlled by two hyperparameters. The first is the \emph{width multiplier} \(\alpha \in (0, 1]\), which controls the number of channels in each layer of the network. Smaller values of \(\alpha\) reduce the number of parameters in each layer of the network uniformly, also reducing computational cost. The second parameter is the \emph{resolution multiplier} \(\rho \in (0, 1]\), which is used to reduce the resolution of the input images and, consequently, the number of operations throughout all layers of the network.

Additional features introduced in MobileNetV1 are batch normalization~\cite{Ioffe2015} for learning stabilization, as well as depthwise-separable convolutions~\cite{Sifre2014}, a convolution operation that uses fewer parameters to achieve comparable results to regular convolutions.

MobileNetV2~\cite{Sandler2018} advanced the state-of-the-art by introducing linear bottleneck layers in the network, reducing the size of the inputs in subsequent layers while preventing information from being lost by non-linear activation functions. The ReLU6 non-linearity (whose first appearance was tracked down to~\cite{Krizhevsky2010}) was chosen instead of regular ReLU to prevent loss of information when calculations with low-precision data types are performed.

Finally, MobileNetV3~\cite{Howard2019a} employs multiple neural architecture search (NAS) algorithms to optimize network architectures for different types of hardware, followed by the manual simplification of the most computationally expensive parts of the generated models. This network employs a variant of the Swish non-linearity~\cite{Ramachandran2017} called h-swish.

\subsection{Single-Shot MultiBox Detector}

The Single-shot MultiBox Detector (SSD)~\cite{Liu2016a} is a technique that utilizes a convolutional neural network, called the base network, combined with multiple subsequent convolutional filters of different sizes, to perform detection under different scales and aspect ratios in multiple regions of an input image. The feature maps of the base network may be pretrained in a classification or detection class. When training the network for a detection task, SSD employs techniques such as data augmentation and hard negative mining for faster training, as well as a loss function that is a weighted sum of both localization and classification losses.

\subsection{You Only Look Once}

You Only Look Once (YOLO)~\cite{Redmon2015} simplified the object detection problem, which was then composed of a region proposal step followed by an image classification step, to a single regression step composed of bounding box coordinates and class probabilities. YOLOv2~\cite{Redmon2016} introduced the use of anchor boxes to the algorithm, a technique that allowed the detection of multiple objects with different aspect ratios in the same quadrant of an input image, using only convolutional layers.

YOLOv3~\cite{Redmon2018} introduces the prediction of bounding box coordinates across multiple scales and the use of residual layers~\cite{He2016} to speed up training, while YOLOv4~\cite{Bochkovskiy2020} adapts multiple data augmentation and feature extraction techniques to allow efficient training and inference of a model on a single GPU with 8 to 16 GB of VRAM.

\section{Related Work}
\label{sec:related}

This section presents recent works that attempt to detect or track soccer balls and other objects relevant to the humanoid soccer scenario.

In~\cite{Poppinga2019}, a model called JET-Net is proposed for the task of detecting robots and soccer balls in a humanoid soccer environment for NAO soccer games. The network uses building blocks popularized in the MobileNets, such as depthwise separable convolutions, as well as working with images in a single channel in order to reduce data complexity, achieving an inference time of 9 ms per frame. % usa focal loss

In~\cite{Szemenyei2019}, ROBO is presented, a network inspired by TinyYOLOv3 which claims to achieve higher accuracy while being 35 times faster. This is achieved by reducing the topology of TinyYOLOv3 further, given that a reduced number of classes will be detected during the robot soccer task; adapting the input resolution of the network to that of the NAO camera; performing downsampling in the input image in the first layers of the network; and using a single anchor box for each class, since the overall shape of object classes is predictable.

A two-stage ball detection algorithm is presented in~\cite{Teimouri2019}, where region proposals are generated first and then passed to a convolutional network for soccer ball detection, using SSD. The authors report an inference time of 5.13 ms in an Intel NUC with a Core i3 CPU.

\emph{DeepBall}~\cite{Komorowski2019} is a recent fully-convolutional neural network architecture inspired by SSD and YOLO, two techniques covered in this work. DeepBall was created to detect soccer balls in long shot videos of real soccer games.

In~\cite{Kukleva2019}, neural networks employing temporal convolutions (TCN), ConvLSTM and ConvGRU layers are used to detect and track the movement of soccer balls in a video feed from a humanoid robot. These networks are trained in sequences of images and the authors report the challenge of gathering sequential data to train the network, resorting to synthetic data for pretraining. A final inference time of around 6 ms is reported for TCN networks using an NVIDIA GTX 1050 Ti.

\section{Methodology}
\label{sec:methodology}

The final goal of this work is to develop a computer vision system for an autonomous humanoid robot, allowing the robot to detect a soccer ball in a compatible time with the dynamics of the game. The vision system may also run on robots with embedded computers, such as mini-PCs, which are generally CPU-only. To select the object detection technique and underlying neural network that compose the vision system, this paper presents a study that investigates the trade-off between accuracy and inference time of multiple neural network architectures proposed specifically for mobile or embedded hardware settings, with different configurations, when executed on this kind of computer.

\subsection{Network architectures and training}

Twenty MobileNetV2 configurations were tested by modifying the values for the width and resolution multipliers. For the width multiplier, the values $1$, $0.75$, $0.5$ and $0.35$ were used, with the resulting networks having $3.47$, $2.61$, $1.95$ and $1.66$ million trainable parameters, respectively.

The values used for the resolution multiplier were chosen so that the input resolution of the network is equal to $224$, $192$, $160$, $128$ and $96$. The combined values of both hyperparameters resulted in a total of twenty models that were trained using the soccer ball data set, described on section \ref{sec:dataset}.

MobileNetV3 models are composed of the ``Large'' and ``Small'' variants presented in~\cite{Howard2019a}, both with width multipliers of $1$ and $0.75$, possessing 5.4 (Large, \(\alpha=1\)), 4 (Large, \(\alpha=0.75\)), 2.9 (Small, \(\alpha=1\)) and 2.4 million (Small, \(\alpha=0.75\)) trainable parameters, as well as minimalistic versions of both variants with \(\alpha=1\), possessing 3.9 and 2 million parameters. Minimalistic models do not contain the more advanced squeeze-and-excite units, hard-swish, and $5 \times 5$ convolutions operations from the non-minimalistic counterparts. YOLO models are composed of the v3 and v4 versions of the neural networks, presented in~\cite{Redmon2018} and~\cite{Bochkovskiy2020} respectively, as well as their ``tiny'' counterparts.

All models used pretrained weights learned in the COCO dataset~\cite{Lin2014}.\footnote{In order to facilitate the replication of these results and encourage the development of similar approaches by other researchers, the software used in this paper is available for download at: \url{https://github.com/douglasrizzo/JINT2020-ball-detection}.}

\subsection{Training procedure}

The models were trained in a server with Intel Xeon Gold 5118@2.3 GHz processors totaling 48 CPUs, 192 GB of RAM and an NVIDIA Tesla V100-PCIE with 16 GB of memory, running CentOS 7.6.1810. The MobileNet models were trained for 50000 training steps. The YOLO and TinyYOLO models were trained for a total of 6000 training steps, following recommendations from the original developers of the model, given the number of classes to be detected.

All MobileNetV2 models were trained using batches of 32 images and the RMSProp optimizer with initial learning rate of $4 \cdot 10^{-3}$, an exponential decay schedule with a decay factor of 0.95 and a momentum coefficient of $0.9$. All MobileNetV3 models were trained using batches of 32 images, stochastic gradient descent with initial learning rate of $0.4$, a cosine decay schedule and a momentum coefficient of $0.9$.

YOLO and TinyYOLO models were trained using batches of 64 images, stochastic gradient descent with a momentum coefficient of 0.9. YOLOv3 and TinyYOLOv3 models used a learning rate of $10^{-3}$, while YOLOv4 and TinyYOLOv4 used a learning rate of $2 \cdot 10^{-3}$. Two step decays at 80\% and 90\% of the training were applied to these learning rates.

\subsection{Humanoid Robot}
\label{sec:roboto}

\begin{figure}
	\centering
	\subfloat[One of the teen-sized robots of the RoboFEI team\label{fig:robot}]{\includegraphics[width=0.35\columnwidth]{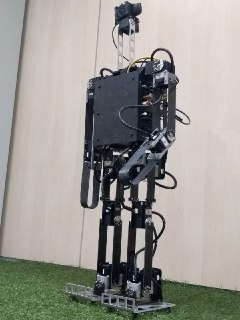}}
	\subfloat[Examples of images from the soccer ball data set\label{fig:dataset}]{\includegraphics[width=.5\columnwidth]{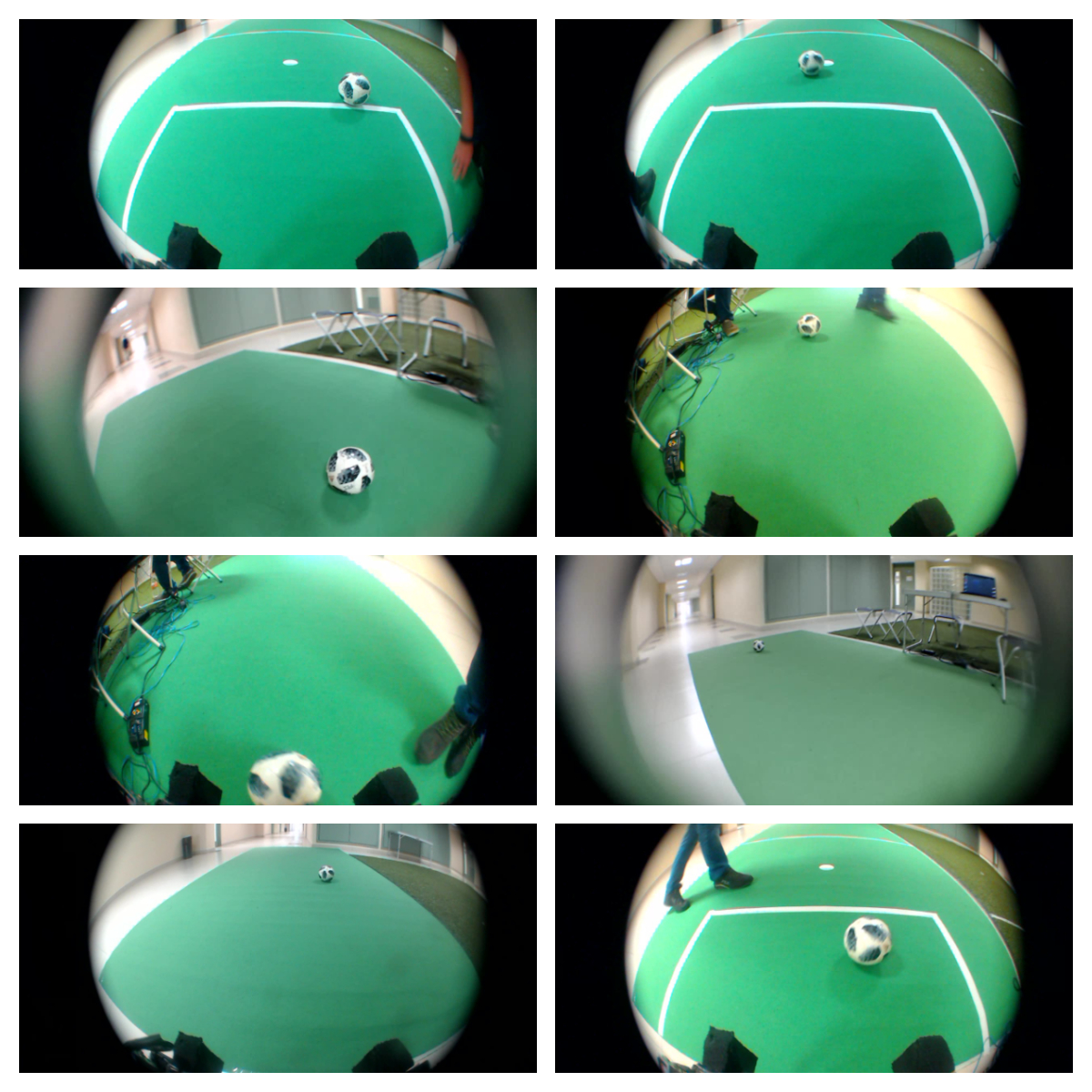}}
	\caption{Image data set and humanoid robot used to collect it.}
	\label{fig:robo-dataset}
\end{figure}

The humanoid robot to which the object detection system is geared towards weighs about 5.9 kg and measures 81 cm in height. It is composed of 19 Dynamixel servomotors (a combination of MX-64, MX-106 and XM430 models), totaling 19 degrees of freedom. The humanoid robot uses a Genius WideCam F100 (Full HD) camera for image capture and a CH Robotics UM7 orientation sensor. The center of mass has a height of 36.1 cm and the robot has a foot area of 174 \(cm ^ 2\). Other measurements include 39.5 cm of shoulder length, 38.5 cm of leg height, 18.8 cm of neck height and 38.5 cm of arm length. The robot is equipped with an Intel NUC Core i7 mini-PC. A picture of the robot is presented in figure \ref{fig:robot}.

\subsection{Image data set}
\label{sec:dataset}

The data set used in this work~\cite{Bianchi2020}\footnote{The dataset was also made available at \url{http://ieee-dataport.org/open-access/open-soccer-ball-dataset}} consists of 4364 images in \(1920 \times 1080\) resolution, collected from the point-of-view of the humanoid robot described in section~\ref{sec:roboto}. A fish-eye lens is used to maximize the field of view of the robot and so all images in the data set also inherit this feature.

Of the 4364 images, 4014 compose the annotated training set and 250, the annotated test set. In these sets, the soccer balls visualized by the robot have been marked with bounding boxes. The training and test sets were collected from different sets of videos, with the purpose of minimizing data correlation.

Each image contains a single soccer ball, captured under multiples lighting conditions, as well as at different angles and distances from the camera. There are pictures of both stationary and moving soccer balls. Figure ~\ref{fig:dataset} presents examples of the data set.

\section{Experiments}
\label{sec:experiments}

In order to recommend the best neural network for the soccer ball detection system, we trained multiple MobileNet and YOLO models in the training set of the data described in section \ref{sec:dataset} and measured the mean average precision (mAP) in the test set during training, taking the final mAP at the end of training as well as the \emph{de facto} measure of precision for all models. Another factor measured by the experiments was inference time, which was measured using a 30-second video taken from the robot's point-of-view in \(1920 \times 1080\) resolution, which was also scaled down to \(1280 \times 720\), \(640 \times 480\) and \(480 \times 360\) pixels. All networks then processed these four versions of the same video and the mean inference time over all frames of each video for each network was recorded.

The MobileNet implementations selected for this work are provided in the TensorFlow Object Detection API~\cite{Huang2017a}, while the YOLO implementations are provided by~\cite{Bochkovskiy2020}. The constrained hardware configuration in which the inference time of the models was captured is equipped with an i5-4210U CPU @ 1.70GHz and 8 GB of RAM and no GPU, in line with the hardware typically used by an autonomous mobile robot. For comparison, the same experiments were performed in the training computer, under GPU and CPU-only settings. In all cases, when CPU-only experiments were executed, all CPU cores were allowed to be utilized.

\section{Results}
\label{sec:results}

Table~\ref{tab:map} displays the mAP of all trained models in the test set, as well as the inference time in milliseconds for different hardware configurations. In this table, the inference time was calculated with the videos in their native \(1920 \times 1080\) resolution. These results allow us to conclude that the canonical YOLOv3 and YOLOv4 implementations, as well as their tiny counterparts, are not optimized for inference on CPUs, achieving the highest inference times of all models in the Intel Core i5-4210U. In fact, it is stated in~\cite{Bochkovskiy2020} that the model implementations are optimized for inference on single GPUs. This can be seen by the comparatively low times achieved in the Tesla V100 GPU, especially by the TinyYOLO models, which achieved the lowest inference times of all models.

\begin{table}
	\centering
	\caption{Average precision (higher is better) and inference time in milliseconds (lower is better) of the trained models. The five best in each category are marked in bold.}
	\label{tab:map}
	\begin{tabular}{r|cccccc}
		                                      & \textbf{Width}        & \textbf{Input} &                   & \multicolumn{3}{c}{\textbf{Inference time (ms)}} \\
		\multicolumn{1}{c|}{\textbf{Network}} & \textbf{mult.}        & \textbf{res.}  & \textbf{mAP}      & \textbf{Core i5} & \textbf{V100}   & \textbf{Xeon} \\ \midrule
		\multirow{20}{*}{MobileNetV2}         & \multirow{5}{*}{0.35} & 96             & 0.4065            & 99.964           & 65.37           & 64.959 \\
		                                      &                       & 128            & 0.7095            & 101.166          & 51.062          & 49.642 \\
		                                      &                       & 160            & 0.6304            & 88.651           & 52.642          & 50.97 \\
		                                      &                       & 192            & 0.6756            & 87.006           & 53.613          & 52.232 \\
		                                      &                       & 224            & 0.4984            & \textbf{78.553}  & 42.852          & \textbf{41.703} \\ \cmidrule{2-7}
		                                      & \multirow{5}{*}{0.5}  & 96             & 0.4065            & 91.403           & 58.919          & 57.626 \\
		                                      &                       & 128            & 0.6986            & 81.08            & 44.529          & 43.388 \\
		                                      &                       & 160            & 0.3361            & 91.775           & 58.288          & 58.616 \\
		                                      &                       & 192            & 0.0944            & 116.076          & 65.629          & 64.828 \\
		                                      &                       & 224            & 0.3253            & \textbf{78.624}  & 42.759          & 43.18 \\ \cmidrule{2-7}
		                                      & \multirow{5}{*}{0.75} & 96             & 0.7284            & 86.569           & 51.065          & 51.528 \\
		                                      &                       & 128            & 0.6954            & 84.159           & \textbf{42.097} & \textbf{41.866} \\
		                                      &                       & 160            & 0.6679            & 81.351           & \textbf{41.883} & \textbf{42.309} \\
		                                      &                       & 192            & 0.6952            & \textbf{78.699}  & \textbf{42.347} & \textbf{41.776} \\
		                                      &                       & 224            & 0.7874            & 85.186           & 48.343          & 47.854 \\ \cmidrule{2-7}
		                                      & \multirow{5}{*}{1}    & 96             & \textbf{0.8133}   & 122.853          & 56.992          & 57.83 \\
		                                      &                       & 128            & 0.7672            & 82.277           & 46.921          & 47.799 \\
		                                      &                       & 160            & \textbf{0.8597}   & 88.886           & 52.278          & 52.569 \\
		                                      &                       & 192            & 0.3632            & 110.75           & 61.263          & 60.052 \\
		                                      &                       & 224            & \textbf{0.8177}   & 79.547           & 42.438          & \textbf{42.183} \\ \midrule
		MobileNetV3 (large min.)              & 1                     & 224            & 0.6007            & 85.808           & 58.706          & 59.581 \\
		MobileNetV3 (large)                   & 0.75                  & 224            & \textbf{0.8847}   & 89.362           & 63.515          & 63.703 \\
		MobileNetV3 (large)                   & 1                     & 224            & 0.6875            & 120.045          & 88.017          & 91.369 \\
		MobileNetV3 (small min.)              & 1                     & 224            & 0.6024            & \textbf{79.142}  & 48.68           & 49.236 \\
		MobileNetV3 (small)                   & 0.75                  & 224            & 0.7067            & \textbf{60.654}  & 49.328          & 47.741 \\
		MobileNetV3 (small)                   & 1                     & 224            & \textbf{0.8651}   & 96.975           & 70.689          & 70.017 \\ \midrule
		TinyYOLOv3                            &                       &                & 0.3381            & 588.235          & \textbf{33.557} & 85.47 \\
		TinyYOLOv4                            &                       &                & 0.3504            & 714.286          & \textbf{29.851} & 119.048 \\
		YOLOv3                                &                       &                & 0.1355            & 5000             & 44.248          & 588.235 \\
		YOLOv4                                &                       &                & 0.1419            & 5000             & 50              & 833.333
	\end{tabular}
\end{table}

As for the MobileNet models, we can see that MobileNetV3 and MobileNetV2 with \(\alpha=1\) achieved the highest mAP in the test data set. However, with the high variation in inference times between all combinations of hyperparameters, the results from Table~\ref{tab:map} alone do not provide enough information to compare model performances. To remedy that, we calculate a performance score for each neural network in each hardware setting \(p_{m,h} = \frac{mAP_m}{t_{m,h}}\), where \(mAP_m\) represents the mAP of network \(m\) (hardware independent) and \(h\), the hardware setting the inference time \(t\) of model \(m\) was gathered from. Then, the performance scores of all models in the same hardware are normalized by the highest performance score in that hardware, leading to the normalized score \(s_{m, h} = \frac{p_{m,h}}{\max_{\eta} p_{\eta,h}}\). Achieving a normalized score \(s_{m, h} = 1\) means that model \(m\) had the highest mAP/inference time ratio off all models in that hardware.\footnote{This score may easily break or be less informative if one neural network in the sample has disproportionately low inference time or high mAP. However, given the well-behaved values presented in Table~\ref{tab:map}, we consider the use of the proposed score appropriate for the purposes of our analysis.}

Table~\ref{tab:scores} presents the normalized scores of all models. Overall, MobileNetV2 models with width multipliers \(\alpha \in \{0.75, 1\}\) had the best scores of all MobileNetV2 models in all hardware settings. We can also see that the five best models in unconstrained hardware settings are the same for both CPU and GPU. However, when operating under the Intel Core i5 4210U processor, both MobileNetV3 models (large and small) with \(\alpha = 0.75\) achieved the highest scores, making MobileNetV3 models a viable option for an object detection system that operates under constrained hardware settings, whereas they did not exhibit the same performance in GPUs.

\begin{table}
	\centering
	\caption{Normalized scores of the trained models under different hardware. Higher is better. A normalized score of 1 indicates that the model performed the best under that hardware, compared with the others.}
	\label{tab:scores}
	\begin{tabular}{r|ccccc}
		                                     & \textbf{Width}        & \textbf{Input} & \multicolumn{3}{c}{\textbf{Normalized score}} \\
		\multicolumn{1}{c}{\textbf{Network}} & \textbf{mult.}        & \textbf{res.}  & \textbf{Core i5} & \textbf{V100}  & \textbf{Xeon} \\ \midrule
		\multirow{20}{*}{MobileNetV2}        & \multirow{5}{*}{0.35} & 96             & 0.349            & 0.323          & 0.323 \\
		                                     &                       & 128            & 0.602            & 0.721          & 0.737 \\
		                                     &                       & 160            & 0.61             & 0.622          & 0.638 \\
		                                     &                       & 192            & 0.666            & 0.654          & 0.667 \\
		                                     &                       & 224            & 0.545            & 0.604          & 0.617 \\ \cmidrule{2-6}
		                                     & \multirow{5}{*}{0.5}  & 96             & 0.382            & 0.358          & 0.364 \\
		                                     &                       & 128            & 0.74             & 0.814          & 0.831 \\
		                                     &                       & 160            & 0.314            & 0.299          & 0.296 \\
		                                     &                       & 192            & 0.07             & 0.075          & 0.075 \\
		                                     &                       & 224            & 0.355            & 0.395          & 0.389 \\ \cmidrule{2-6}
		                                     & \multirow{5}{*}{0.75} & 96             & 0.722            & 0.74           & 0.729 \\
		                                     &                       & 128            & 0.709            & \textbf{0.857} & \textbf{0.857} \\
		                                     &                       & 160            & 0.705            & 0.828          & 0.814 \\
		                                     &                       & 192            & 0.758            & \textbf{0.852} & \textbf{0.858} \\
		                                     &                       & 224            & 0.793            & 0.845          & 0.849 \\ \cmidrule{2-6}
		                                     & \multirow{5}{*}{1}    & 96             & 0.568            & 0.741          & 0.726 \\
		                                     &                       & 128            & \textbf{0.8}     & \textbf{0.849} & \textbf{0.828} \\
		                                     &                       & 160            & \textbf{0.83}    & \textbf{0.853} & \textbf{0.844} \\
		                                     &                       & 192            & 0.281            & 0.308          & 0.312 \\
		                                     &                       & 224            & \textbf{0.882}   & \textbf{1}     & \textbf{1} \\ \midrule
		MobileNetV3 (large min.)             & 1                     & 224            & 0.601            & 0.531          & 0.52 \\
		MobileNetV3 (large)                  & 0.75                  & 224            & \textbf{0.85}    & 0.723          & 0.716 \\
		MobileNetV3 (large)                  & 1                     & 224            & 0.492            & 0.405          & 0.388 \\
		MobileNetV3 (small min.)             & 1                     & 224            & 0.653            & 0.642          & 0.631 \\
		MobileNetV3 (small)                  & 0.75                  & 224            & \textbf{1}       & 0.744          & 0.764 \\
		MobileNetV3 (small)                  & 1                     & 224            & 0.766            & 0.635          & 0.637 \\ \midrule
		TinyYOLOv3                           &                       &                & 0.049            & 0.523          & 0.204 \\
		TinyYOLOv4                           &                       &                & 0.042            & 0.609          & 0.152 \\
		YOLOv3                               &                       &                & 0.002            & 0.159          & 0.012 \\
		YOLOv4                               &                       &                & 0.002            & 0.147          & 0.009
	\end{tabular}
\end{table}

\subsection{Performance under different input resolutions}

This section presents the inference times in milliseconds of all model implementations when processing input videos of multiple resolutions (\(1920 \times 1080\), \(1280 \times 720\), \(640 \times 480\), \(480 \times 360\)). Table~\ref{tab:yolo-inference-time} presents the mean and standard deviation of the inference time of all YOLO models for the tested hardware configurations. Overall, all YOLO and TinyYOLO models achieved low standard deviation in this test, implying that their implementation~\cite{Bochkovskiy2020} is indifferent to the input resolution of images.

\begin{table}
	\centering
	\caption{Inference time of YOLO and TinyYOLO models when processing videos of multiple input resolutions.}
	\label{tab:yolo-inference-time}
	\begin{tabular}{c|rcc}
		                            &                                       & \multicolumn{2}{c}{\textbf{Inference time}} \\
		\textbf{Network}            & \multicolumn{1}{l}{\textbf{Hardware}} & \textbf{mean} & \textbf{std. dev.} \\\midrule
		\multirow{3}{*}{TinyYOLOv3} & Tesla V100                            & 41.957        & 10.022 \\
		                            & Xeon Gold 5118                        & 88.983        & 3.369 \\
		                            & i5-4210U                              & 588.235       & 0.000 \\\midrule
		\multirow{3}{*}{TinyYOLOv4} & Tesla V100                            & 38.808        & 9.136 \\
		                            & Xeon Gold 5118                        & 114.529       & 6.094 \\
		                            & i5-4210U                              & 714.286       & 0.000 \\\midrule
		\multirow{3}{*}{YOLOv3}     & Tesla V100                            & 46.726        & 4.081 \\
		                            & Xeon Gold 5118                        & 588.235       & 0.000 \\
		                            & i5-4210U                              & 5000.000      & 0.000 \\\midrule
		\multirow{3}{*}{YOLOv4}     & Tesla V100                            & 50.385        & 0.676 \\
		                            & Xeon Gold 5118                        & 817.308       & 32.051 \\
		                            & i5-4210U                              & 5000.000      & 0.000
	\end{tabular}
\end{table}

As for the MobileNet results, we first discuss the distributions of results in the three hardware settings. In total, 104 values were collected in each setting (twenty V2 and six V3 models applied to videos in four resolutions), with the following statistics: \(\mu_\mathrm{i5}=68.292\), \(\sigma_\mathrm{i5}=17.374\), \(\mu_\mathrm{V100}=47.24\), \(\sigma_\mathrm{V100}=8.756\), \(\mu_\mathrm{Xeon}=47.009\) and \(\sigma_\mathrm{Xeon}=8.981\).

Due to the similarity in the inference times collected from the NVIDIA Tesla V100 GPU and the Intel Xeon Gold 5118, we executed a two-sample Kolmogorov-Smirnov test between the two samples, with a result of \(p=0.97371\), indicating that the measurements collected in the 48 CPUs and the single GPU are similar with high statistical relevance. Because of that, in this section we only report results for the MobileNets in the NVIDIA Tesla V100 GPU and the Intel i5-4210U processor.

The distance between \(\mu_\mathrm{i5}\) and \(\mu_\mathrm{V100}\) is indicative of the performance lost by executing deep learning models in constrained CPUs, while a larger standard deviation on the CPU (\(\sigma_\mathrm{i5}\)) indicate that there is a larger variability in network performance, given the resolution of the input video.

\begin{figure}
	\centering
	\subfloat[MobileNetV2 by input resolution on i5-4210U]{\includegraphics[width=.5\columnwidth]{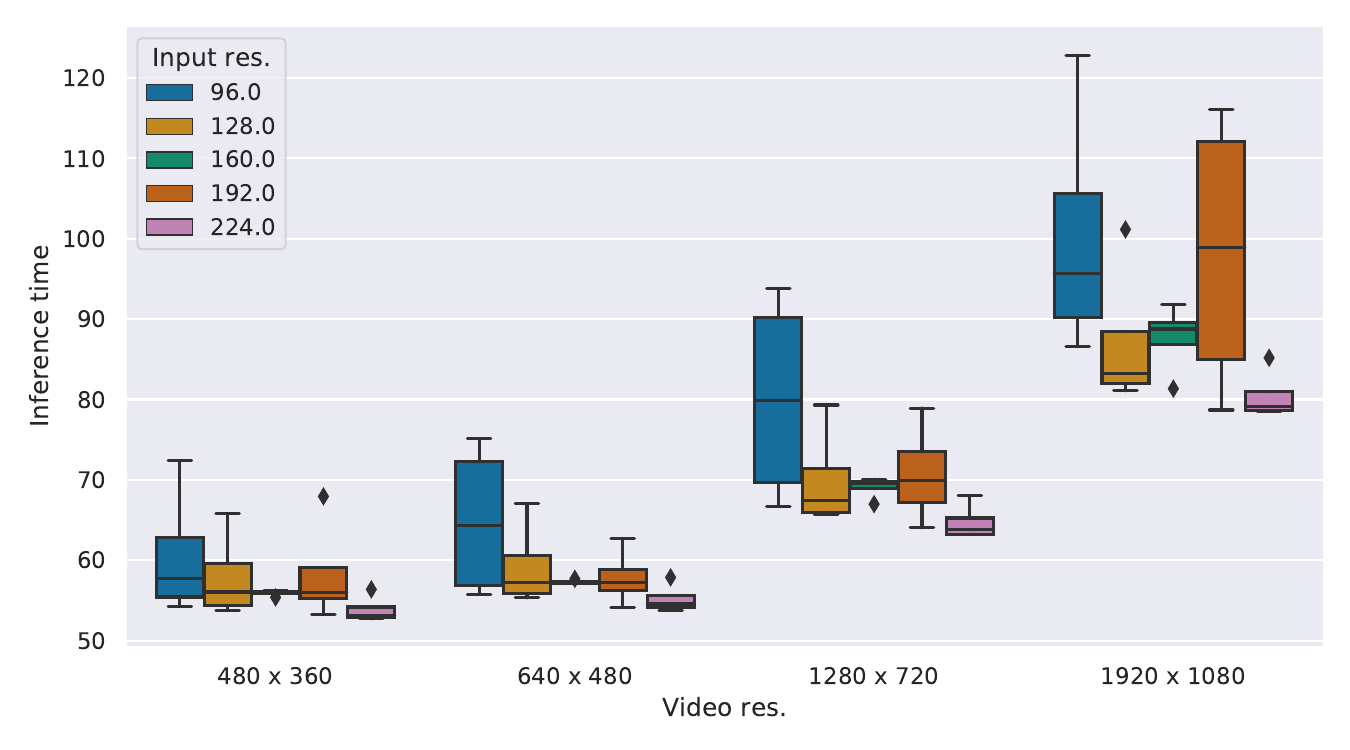}}
	\subfloat[MobileNetV2 by input resolution on V100]{\includegraphics[width=.5\columnwidth]{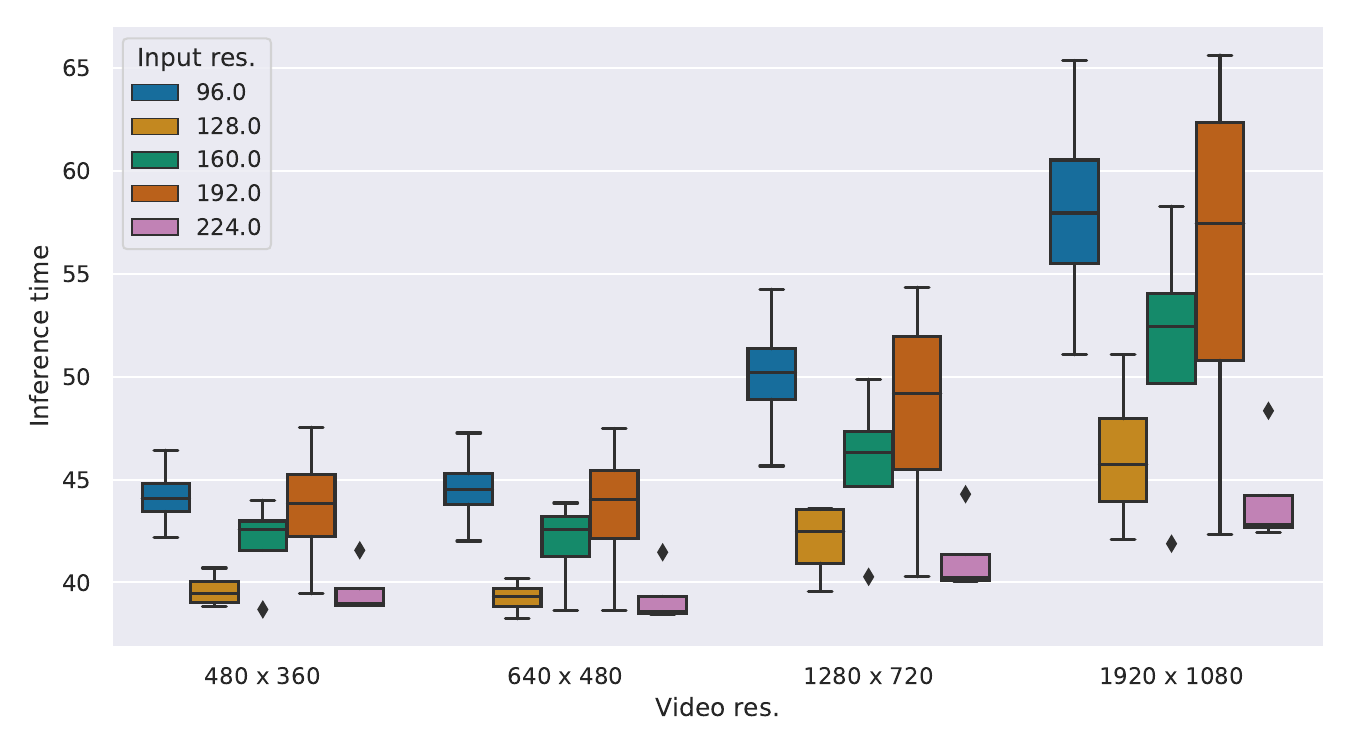}}\\
	\subfloat[MobileNetV2 by width multiplier on i5-4210U]{\includegraphics[width=.5\columnwidth]{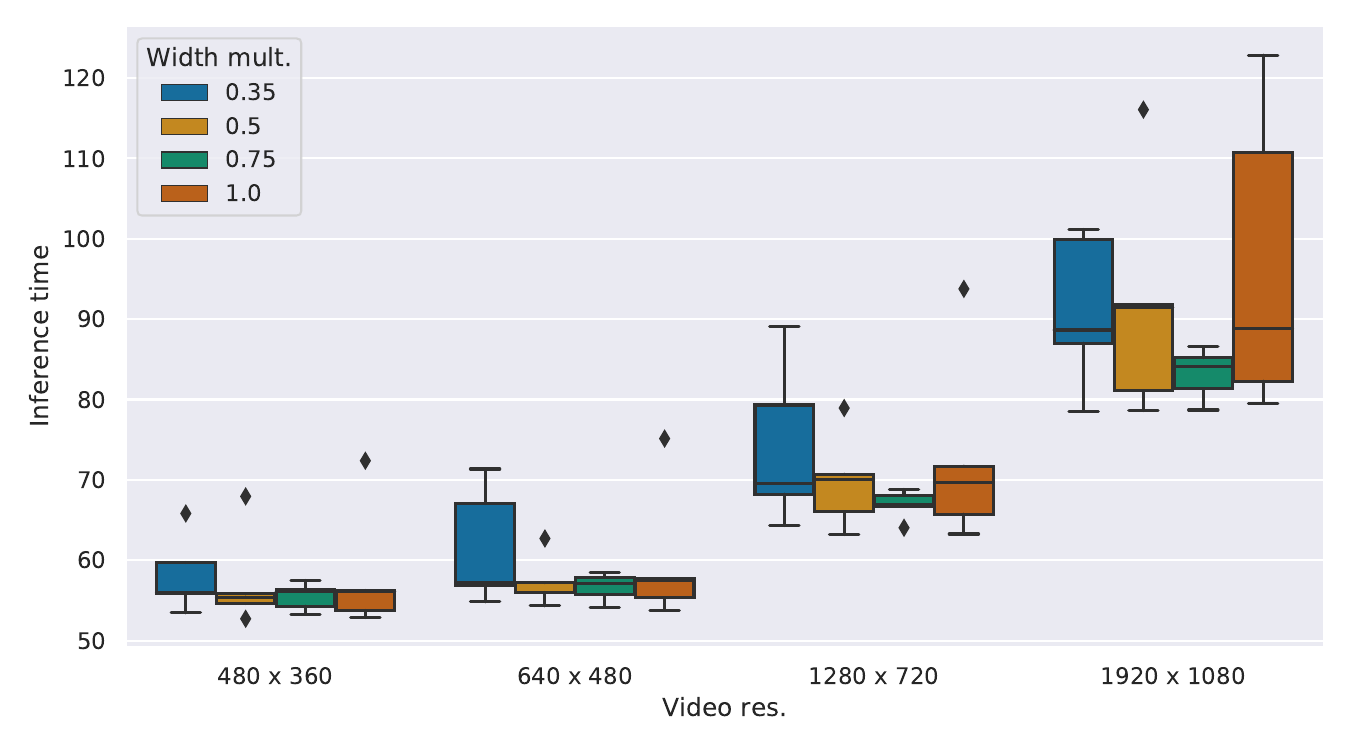}}
	\subfloat[MobileNetV2 by width multiplier on V100]{\includegraphics[width=.5\columnwidth]{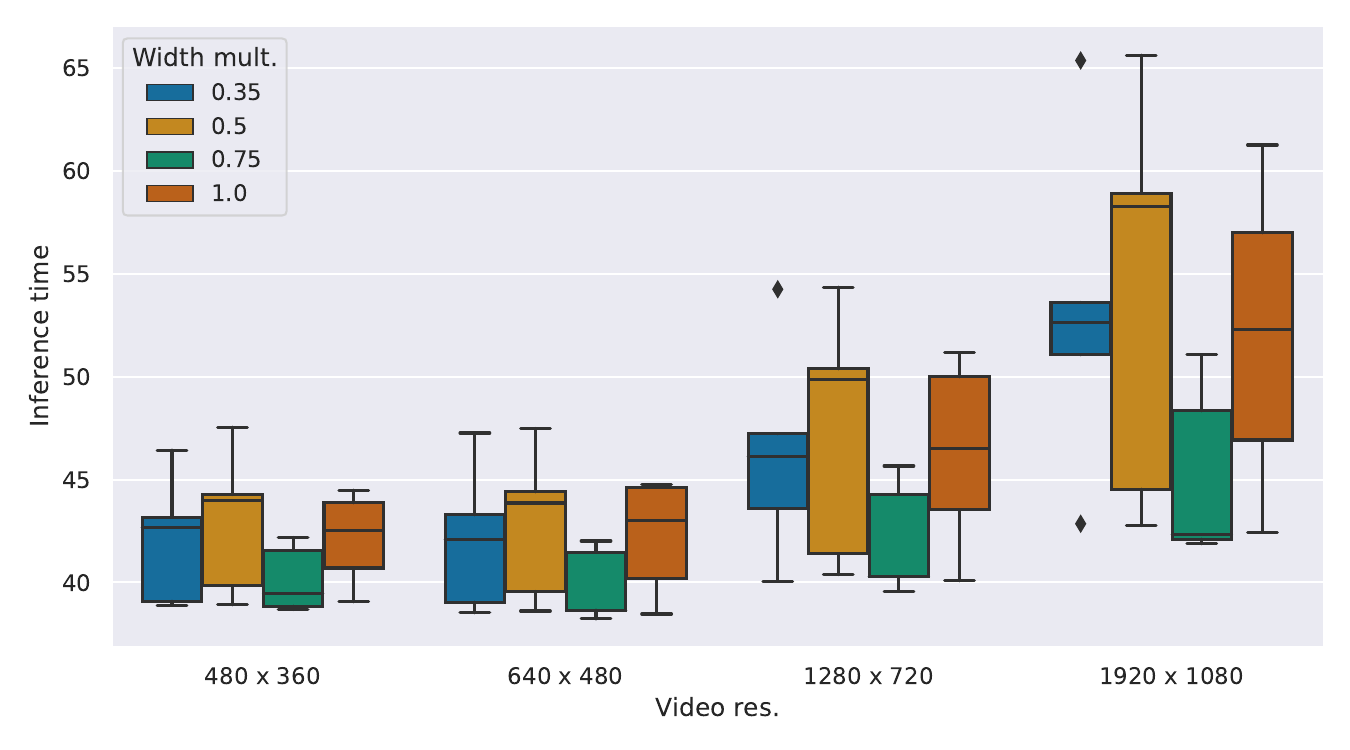}}\\
	\subfloat[MobileNetV3 on i5-4210U]{\includegraphics[width=.5\columnwidth]{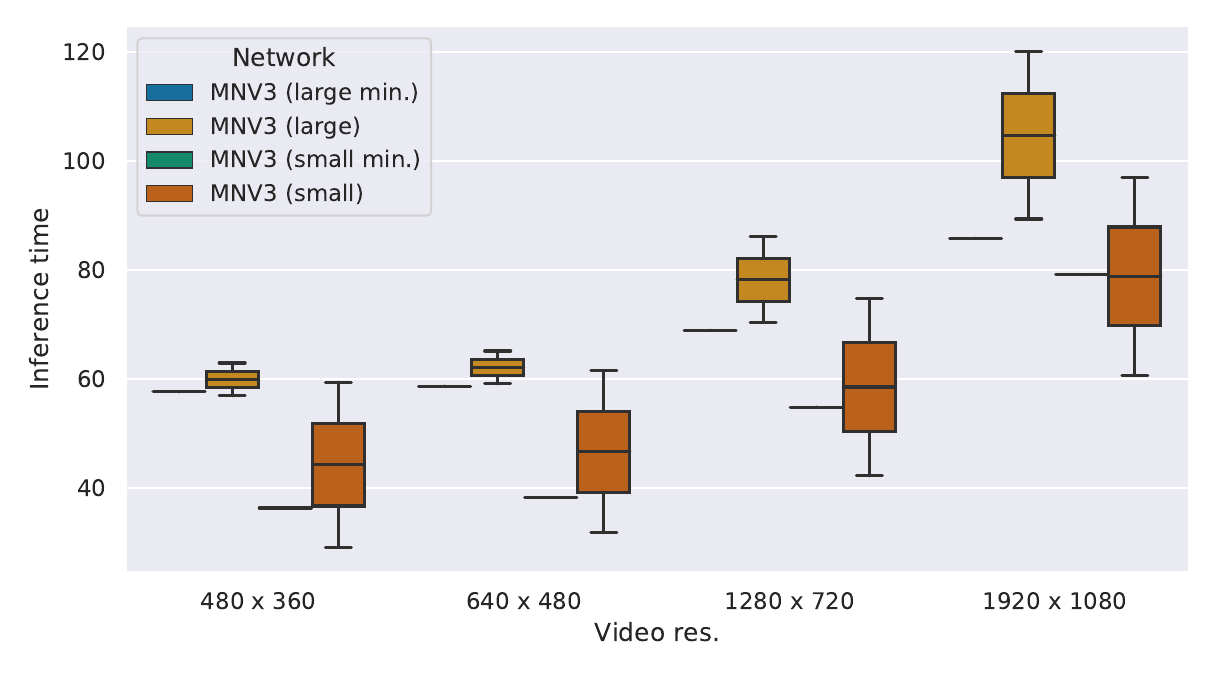}}
	\subfloat[MobileNetV3 on V100]{\includegraphics[width=.5\columnwidth]{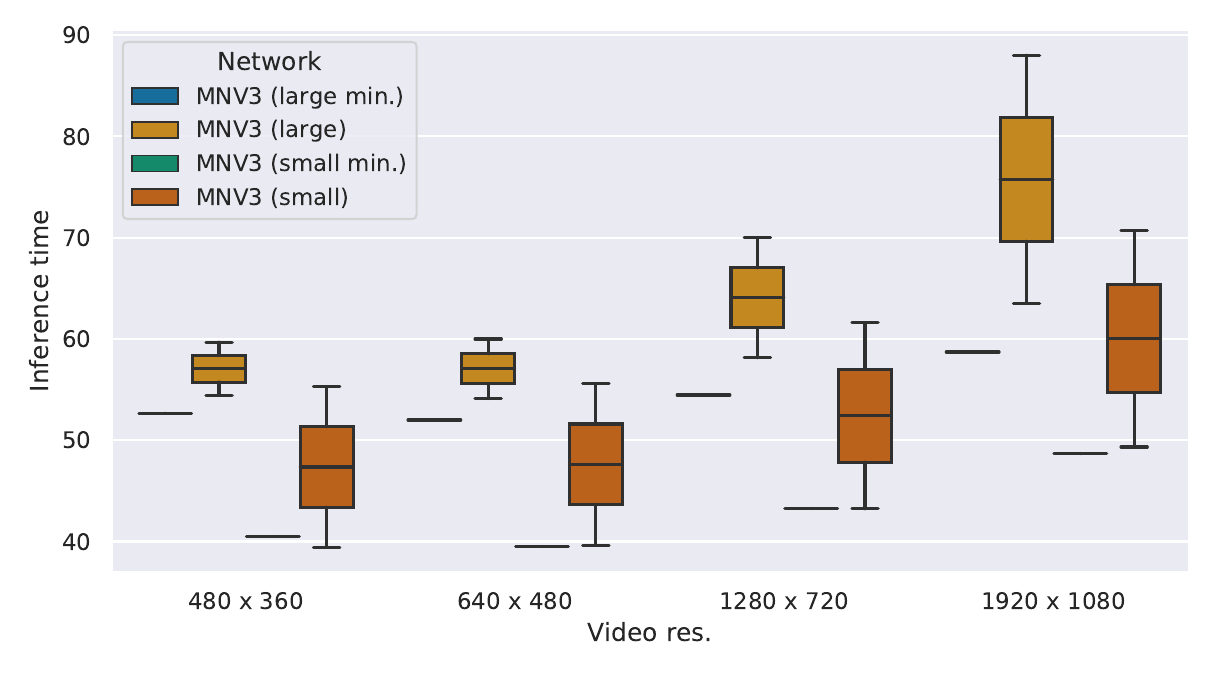}}
	\caption{Inference time of MobileNetV2 and V3 models in Intel i5-4210U (left) and NVIDIA Tesla V100 (right).}
	\label{fig:mobilenets_resolutions}
\end{figure}

Figure~\ref{fig:mobilenets_resolutions} presents the inference time by frame in milliseconds for the MobileNet models when processing the same video under multiple resolutions in the Intel i5-4210U CPU and an NVIDIA Tesla V100 GPU. Unlike the results for the YOLO models, all MobileNets show a significant speedup when applied to input frames of lower resolutions. This information may be relevant, as the implementations by~\cite{Huang2017a} already operate in downscaled images, with a resolution of \(300 \times 300\), indicating that downscaling the input feed prior to executing object detection is a valid strategy to achieve lower inference times. This speedup is visualized in both constrained CPU and GPU settings.

\section{Conclusions}
\label{sec:conclusions}

This work presented a comparative study of the performance of multiple neural network architectures, designed for fast inference with a reduced number of trainable parameters while maintaining high precision, when applied in the task of soccer ball detection under constrained and unconstrained hardware scenarios, as well as when processing input images of different resolutions.

Results have shown that MobileNetV2 models with high width multipliers have the best trade-off between mAP and inference time in unconstrained hardware settings, being suitable when executing inference in remote servers is an option. However, under a local, constrained, CPU-only scenario, MobileNetV3 models have shown the best scores, while not having remarkable performance when operating in a GPU. Lastly, the official implementations of YOLO and TinyYOLO, being optimized for inference in GPUs, displayed poor results in our low-end Intel Core i5-4210U processor. 

In future work, we aim to evaluate the performance of state-of-the-art reduced object detection models in embedded systems with GPUs, seem them as the next step in hardware for mobile robotics.

\section*{Acknowledgements}
The authors acknowledge the S\~ao Paulo Research Foundation (FAPESP Grant 2019/07665-4) for supporting this project. This study was financed in part by the Coordena\c{c}\~ao de Aperfei\c{c}oamento de Pessoal de N\'ivel Superior – Brasil (CAPES) – Finance Code 001. This is a preprint of an article published in the Journal of Intelligent \& Robotic Systems. The final authenticated version is
available online at: https://doi.org/10.1007/s10846-021-01336-y.

\section*{Conflict of interest}
The authors declare that they have no conflict of interest.

\bibliographystyle{unsrt}  

\end{document}